\documentclass[11pt,a4paper]{article}

\usepackage[hyperref]{emnlp-ijcnlp-2019}
\usepackage{times}
\usepackage{latexsym}
\usepackage{array}
\usepackage{booktabs}
\usepackage{diagbox}
\usepackage{url}

\usepackage{multirow}
\usepackage{balance}
\usepackage{amsfonts}
\usepackage{amsmath}
\usepackage{amssymb}
\usepackage{graphicx}
\usepackage{subfigure}
\usepackage{microtype}
\usepackage{xspace}
\usepackage{url}
\usepackage{enumitem}
\usepackage{wrapfig,lipsum}

\aclfinalcopy

\newcommand{\wrong}[1]{\textbf{\textcolor{red}{#1}}}

\def \b {\mathbf{b}}

\def \f {\mathbf{f}}

\def \h {\mathbf{h}}

\def \r {\mathbf{r}}

\def \y {Y}

\def \F {\mathbf{F}}

\def \Y {\mathbf{Y}}
\def \ie {\textit{i.e.}}
\def \eg {\textit{e.g.}}

\usepackage{tabularx, booktabs}
\newcolumntype{Y}{>{\centering\arraybackslash}X}
\newcolumntype{L}{>{\arraybackslash}X}

\newtheorem{thm:def}{Definition}
\newtheorem{thm:eg}{Example}
\newtheorem{thm:lem}{Lemma}
\newtheorem{thm:obs}{Observation}
\newtheorem{thm:req}{Requirement}

\newcommand{\nop}[1]{}

\newcommand{\fl}{{ F$_1$ }}

\newcommand{\method}[1]{\mbox{#1}\xspace}
\newcommand{\our}{\method{Neural-Char-CRF}}

\def \F {\mathcal{F}}

\def \Z {\mathcal{Z}}

\def \Y {Y}

\def \b {\mathbf{b}}
\def \r {\mathbf{r}}
\def \f {\mathbf{f}}
\def \h {\mathbf{h}}

\def \y {y}
\def \z {\mathbf{z}}

\DeclareMathAlphabet{\mathbbold}{U}{bbold}{m}{n}

\newcommand{\smallsection}[1]{\smallskip \noindent\textbf{#1.}}    

\definecolor{themered}{HTML}{FF8375}
\usepackage{xparse}
\NewDocumentCommand{\heng}{ mO{} }{\textcolor{red}{\textsuperscript{\textit{Heng}}\textsf{\textbf{\small[#1]}}}}

\author{
Liyuan Liu$^{\dag}\ $
Zihan Wang$^{\dag}\ $
Jingbo Shang$^{\dag}\ $
Dandong Yin$^{\dag}\ $
Heng Ji$^{\dag}\ $\\
\textbf{
Xiang Ren$^{\sharp}\ $
Shaowen Wang$^{\dag}\ $
Jiawei Han$^{\dag}\ $
}
\\[0.5ex]
{$^{\dag}$ University of Illinois at Urbana-Champaign, Urbana, IL, USA}\\
{$^{\sharp}$ University of Southern California, Los Angeles, CA, USA}\\
{
\begin{footnotesize}
\tt
$^{\dag}$\{ll2, zihanw2, shang7, dyin4, hengji, shaowen
\end{footnotesize}
}
\\
{
\begin{footnotesize}
\tt
hanj\}@illinois.edu
$^{\sharp}$xiangren@usc.edu
\end{footnotesize}
}
}
\date{}

\begin{document}

\title{Raw-to-End Name Entity Recognition in Social Media}
\maketitle


\begin{abstract}
Taking word sequences as the input, typical named entity recognition (NER) models neglect errors from pre-processing (\eg, tokenization).
However, these errors can influence the model performance greatly, especially for noisy texts like tweets. 
Here, we introduce \our, a raw-to-end framework that is more robust to pre-processing errors.
It takes raw character sequences as inputs and makes end-to-end predictions.
Word embedding and contextualized representation models are further tailored to capture textual signals for each character instead of each word.
Our model neither requires the conversion from character sequences to word sequences, nor assumes tokenizer can correctly detect all word boundaries. 
Moreover, we observe our model performance remains unchanged after replacing tokenization with string matching, which demonstrates its potential to be tokenization-free. 
Extensive experimental results on two public datasets demonstrate the superiority of our proposed method over the state of the art.\footnote{The implementations and datasets are made available at: https://github.com/LiyuanLucasLiu/Raw-to-End.}
\end{abstract}

\section{Introduction}
\label{sec:intro}

    \begin{figure}[ht]
    \centering
    \includegraphics[width=\linewidth]{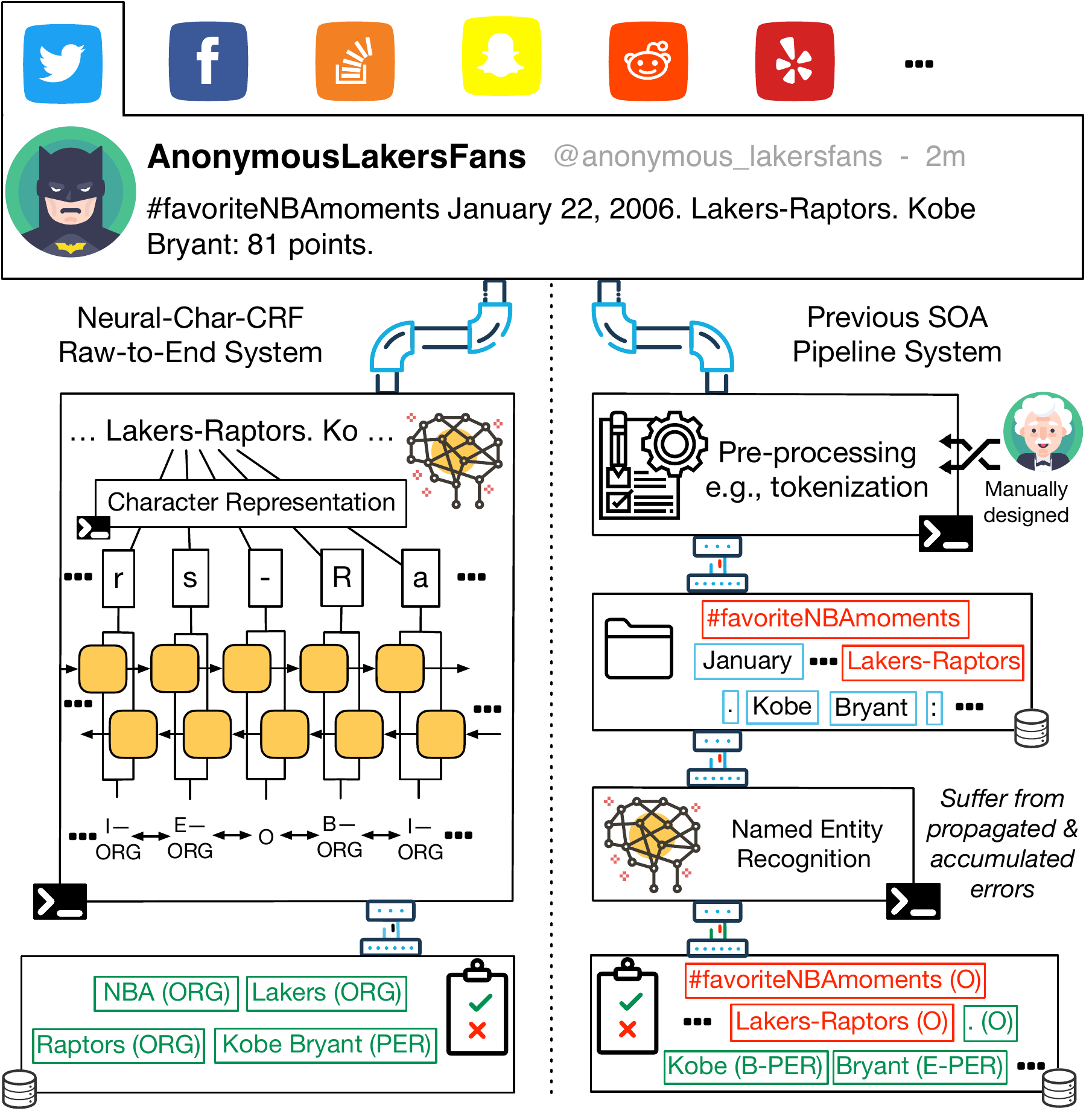}
    \caption{Comparison between Pipeline System and \our. 
    In the pipeline system, pre-processing errors hurt the model performance. 
    The raw-to-end training in \our suffers less from such error propagation and reduces human endeavors.}
    \label{fig:lakers}
\end{figure}


The explosively growing social media generates noisy, irregular texts on a massive scale.
How to digest such fast-evolving text data has become an urgent, challenging topic and attracted a significant amount of attention~\cite{strauss2016results,derczynski2017results}.
Here we focus on one important step towards extracting information from social media posts: named entity recognition (NER).

The key challenge of NER in social media lies in the noisy nature of online posts.
Compared to regular articles (\eg, the Wall Street Journal news), these posts are written in a more informal and creative manner.
Hence, even for pre-processing (\eg, tokenization), it becomes difficult to build tools to analyze such texts with no or little errors. 
Moreover, pre-processing 
is crucial to 
get good NER performance, and thus their errors cannot be neglected. 
For example, on the Broad Twitter Corpus~\cite{derczynski2016broad}, we observe word boundaries of more than 45\% named entities to be incorrectly identified by spaCy~\cite{spacy2}.
At the same time, as in Figure~\ref{fig:lakers}, the pipeline system builds pre-processing and sequence labeling separately, assumes all entity boundaries are correctly identified as word boundaries, and are vulnerable to pre-processing errors. 
For example, when the tokenizer wrongly identifies ``Lakers-Raptors" as one token, the NER model has to make one prediction for the whole string and cannot correctly recognize ``Lakers" and ``Raptors" as two entities.  

In the past few years, many efforts have been made to unify different stages in the pipeline system and these approaches have demonstrated a great potential to reduce human endeavors, alleviate error propagation, and improve the performance. 
For example, neural networks 
achieve good success on
jointly extracting textual signals and detecting entities~\cite{lample2016neural,ma2016end}.
Still, these methods take word sequences as inputs and neglect errors from pre-processing. 
For example, on the BTC dataset, after replacing perfect tokenizer with system tokenizer, the performance drops $24.40$ absolute \fl on average. 

In this paper, we propose \our, a raw-to-end framework that takes raw character sequences as inputs and makes predictions in an end-to-end manner.
It integrates pre-processing with representation learning modules: 
modules are designed to align each character to a pre-trained word embedding. 
Specifically, we introduce two strategies for building this alignment, one leverages tokenizer and the other utilizes string matching.
Besides, we pre-train character-level language models and 
construct contextualized character representations.
Different from the pipeline system, our model captures textual signals for each character instead of each word, predict whether it belongs to an entity, and if so what its position and type are. 
For example, in Figure~\ref{fig:framework}, our model needs to construct representations and make predictions for each character in ``Lakers-Raptors".
Therefore, if the tokenizer wrongly detect the whole string as one token, our model would need to handle noisy representations, but it still has the potential to correctly recognize ``Lakers'' and ``Raptors'' as entities.

We evaluate our model on two public datasets --- Twitter Name Tagging (TNT)~\cite{lu2018visual} and Broad Twitter Corpus (BTC)~\cite{derczynski2016broad}.
Experimental results show the effectiveness of our proposed method, which advances state-of-the-art by $3.70$ and $6.65$ absolute \fl gain on TNT and BTC respectively. 
We also observe that, the performance of our model remains unchanged after replacing tokenization with string matching, which demonstrates the potential of our model to be tokenization-free.
\section{\our}
\label{sec:frame}

In this section, we first introduce the raw-to-end problem formulation, then proceed to the proposed \our framework. 

\subsection{Problem Formulation}

We formulate the raw-to-end NER as a character-level sequence labeling task.
The input is a character sequence $X=\{x_1, x_2, \ldots, x_T\}$, where $x_i$ $(1 \le i \le T)$ is the $i$-th character and $T$ is the input length.
Following the \texttt{IOBES} labeling schemes~\cite{ratinov2009design},
we assign a label for each character.
Specifically, when a sequence of characters is identified as a named entity, its starting, middle and ending character are labeled as \texttt{B-}, \texttt{I-}, and \texttt{E-} respectively (if the entity has only one character, it is labeled as \texttt{S-}), followed by the type; otherwise, it would be labeled as \texttt{O} instead.
Referring the label for $x_i$ as $\y_i$, the goal of the raw-to-end NER is to predict the label sequence $\Y = \{\y_1, \y_2, \ldots \y_T\}$.

\subsection{Framework Architecture}

\begin{figure}[t]
  \centering
  \includegraphics[width=\linewidth]{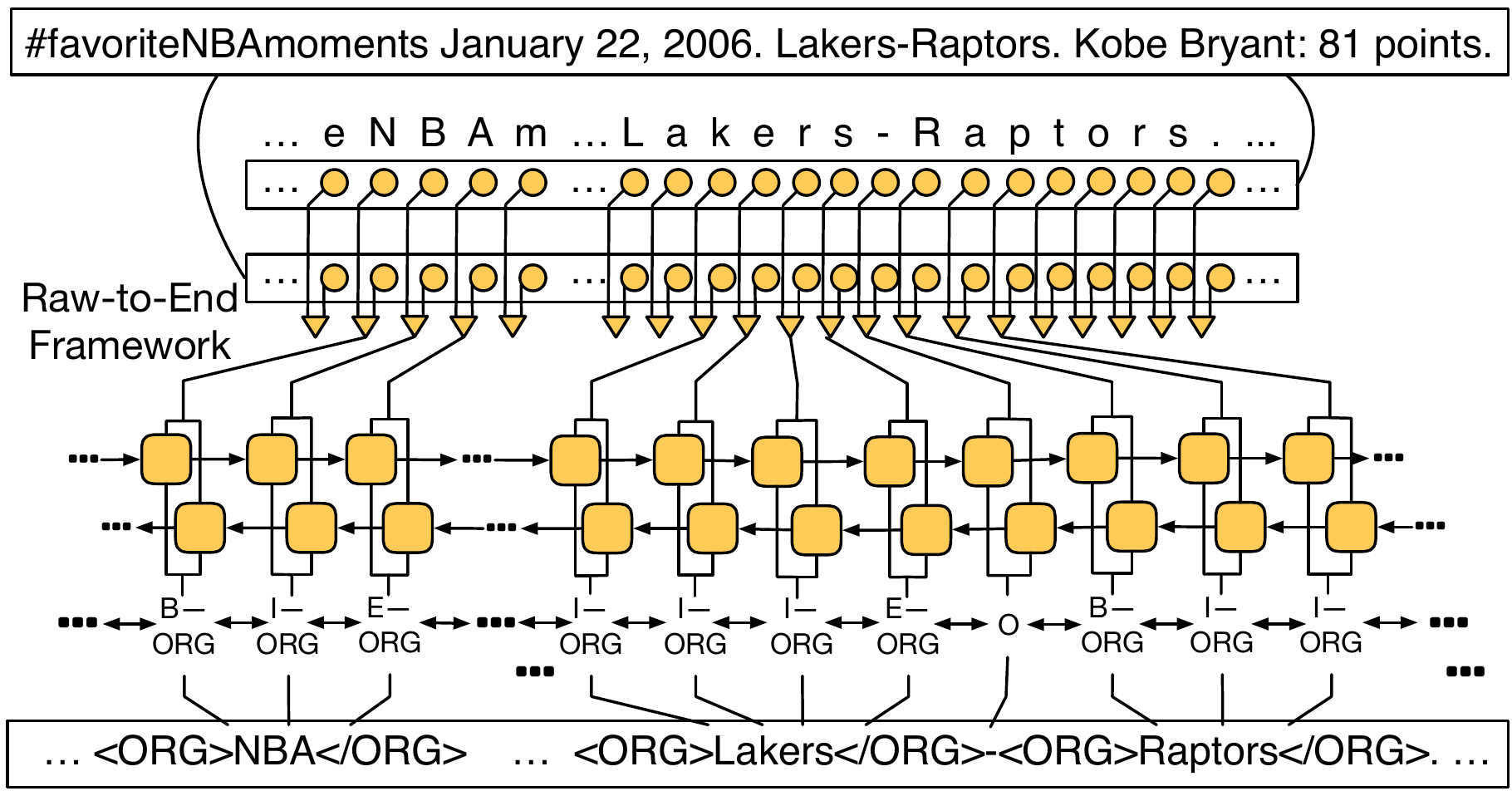}
  \caption{The proposed \our Framework. It accepts the raw text as the input and makes predictions at the character level.}
  \label{fig:framework}
  \vspace{-0.5cm}
\end{figure}

We propose a novel raw-to-end framework, \our, to reduce the error propagation and the reliance on pre-processing. 
This framework directly takes character sequences as inputs and makes prediction for each character.
As visualized in Figure~\ref{fig:framework}, \our first constructs representations for each character, then detects entities with LSTM-CRF. 

\smallsection{Character Representation}
Recent state-of-the-art NER models usually unify representations learned by multiple methods.
We assume there are $n$ different representation modules, namely $M_i$ $1 \le 1 \le n$ (\eg, different pre-trained word embedding or contextualized representation models). 
Given the $j$-th character in the input sequence, the representation vector produced by module $M_i$ is denoted as $\f_{i, j}$.
In this paper, we concatenate the output from different modules as the final representation, \ie, $\f_j = [\f_{1, j}; \f_{2, j}; \cdots; \f_{n, j}]$.
Given the input sequence $X$, we define its representation sequence as $\F = \{\f_1, \f_2, \cdots, \f_T\}$.

\smallsection{Decoding with LSTM-CRF}
Building upon the character representation of the input sequence, we use LSTM-CRF~\cite{huang2015bidirectional} to extract entities: we first feed $\F$ into Bi-LSTMs, whose output is marked as $\Z = \{\z_1, \z_2, \cdots, \z_T\}$.
A linear-chain CRF is further leveraged to model the whole label sequence simultaneously. 
Specifically, for the input sequence $\Z$, CRF defines the conditional probability of $\Y = \{y_1, \cdots, y_T\}$ as 
\begin{equation}
p(\Y|\Z) = \frac{\prod_{t=1}^{T} \phi(y_{t-1}, y_t, \z_t)}{\sum_{\hat{\Y} \in \mathcal{\Y}(\Z)} \prod_{t=1}^{T} \phi(\hat{y}_{t-1}, \hat{y}_t, \z_t)}
\label{eqn:crf}
\end{equation}
where $\hat{\Y} = \{\hat{y}_1, \cdots, \hat{y}_T\}$ is a possible label sequence, $\mathcal{\Y}(\Z)$ refers to the set of all possible label sequences for $\Z$, and $\phi(y_{t-1}, y_t, \z_t)$ is the potential function of the CRF. In this paper, we define the potential function as:
$
\phi(y_{t-1}, y_t, \z_t) = \exp(W_{y_t} \z_t + b_{y_{t-1}, y_t})
$
where $W_{y_t}$ and $b_{y_{t-1}, y_t}$ are the weight and bias.

During the model training, we use the negative log-likelihood of Equation~\ref{eqn:crf} as the loss function. 
In the inference stage, we find the predicted label sequence for input $X$ by maximizing the probability in Equation~\ref{eqn:crf}.
Although the denominator in Equation~\ref{eqn:crf} contains a number of terms exponential to the length $T$, due to the definition of the potential function, both training and inference can be efficiently conducted using dynamic programming.


\begin{figure}[t]
  \centering
  \includegraphics[width=\linewidth]{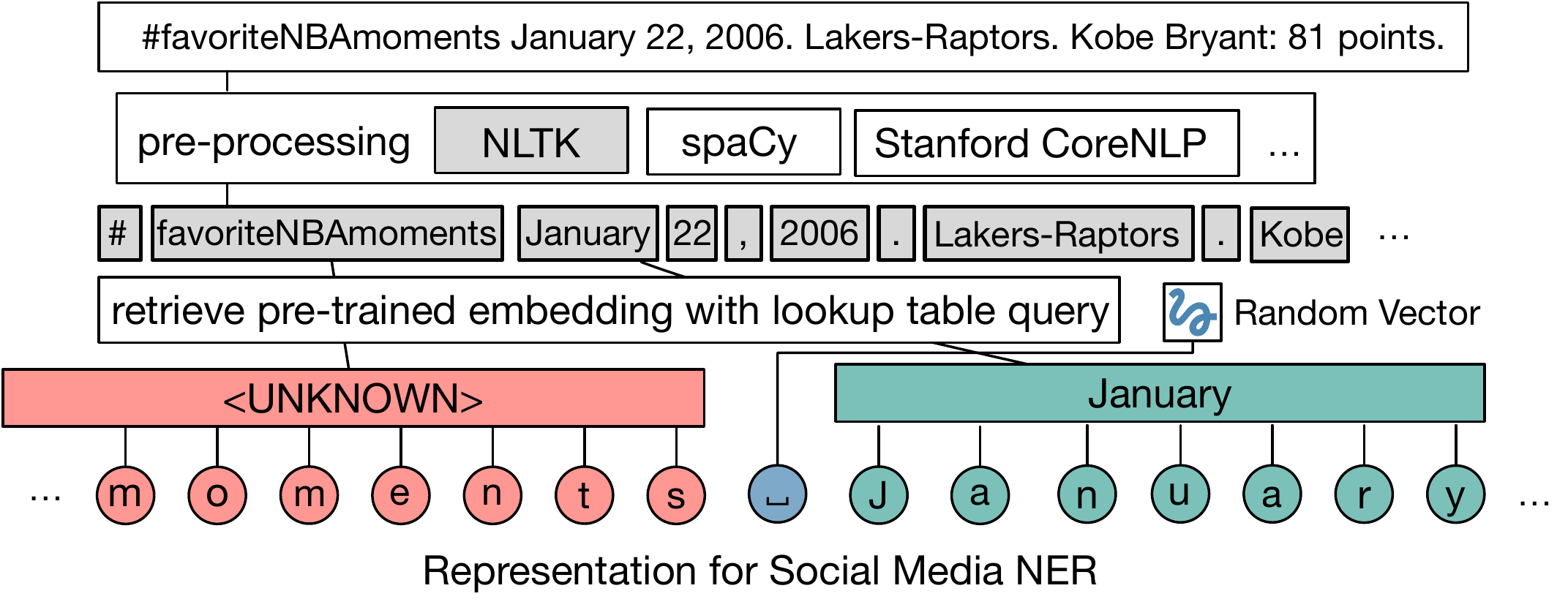}
  \vspace{-0.7cm}
  \caption{Pre-processing tools are leveraged to integrate pre-trained word embedding in a robust manner.}
  \label{fig:embed}
  \vspace{-0.5cm}
\end{figure}

\section{Character Representation Modules}
\label{sec:rep}

In this section, we discuss character representation modules used in \our.
Representation learning techniques at the word level, such as word embedding~\cite{mikolov2013efficient} and contextualized representations~\cite{peters2018deep,devlin2018bert,akbik2018contextual}, have demonstrated their effectiveness in NER.
However, learning representation at the character level has not been explored sufficiently.
We keep the same spirits of word embedding and contextualized representations, and further tailor these two techniques from the word level to the character level.

\subsection{Robust Word Embedding Alignment}

Existing work~\cite{cherry2015unreasonable} has demonstrated that word embedding trained on a web-scale corpus is incredibly powerful in improving Twitter NER performance.
The key reason is that such trained embedding helps the model to better understand rare words in NER datasets.
Trained on web-scale corpus, rare words will appear for a reasonably large number of times, and thus have informative embedding vectors.
Since named entitie often contain rare and uncommon words, such embeddings could provide valuable information for NER.
Therefore, we aim to incorporate word embedding into the character representation.
We align each character to a word, which is specified by tokenization or string matching, then integrate the word embedding into the representation of this character.
In this way, the NER model is aware of word-level signals while not forced to make predictions at word level.
We elaborate these two approaches below.

\smallsection{Tokenization-based Embedding Alignment}
The first approach is designed to leverage pre-processing in a robust manner.
It uses tokenization results to align word embedding down to each character. 
The workflow is visualized in Figure~\ref{fig:embed}.
First, we run a tokenizer to identify the word boundaries and retrieve the pre-trained word embedding with lookup table queries.
Then, for each word, we use its word embedding vector as the representation for all characters that belong to this word.
At the same time, all whitespace characters use the same embedding vector, which is randomly initialized.
In this strategy, the tokenizer is only used in the alignment procedure, the sequence labeling model does not depend on the detected token boundaries and is more robust to error propagation. 

\begin{figure}[t]
  \centering
  \vspace{0.2cm}
  \includegraphics[width=\linewidth]{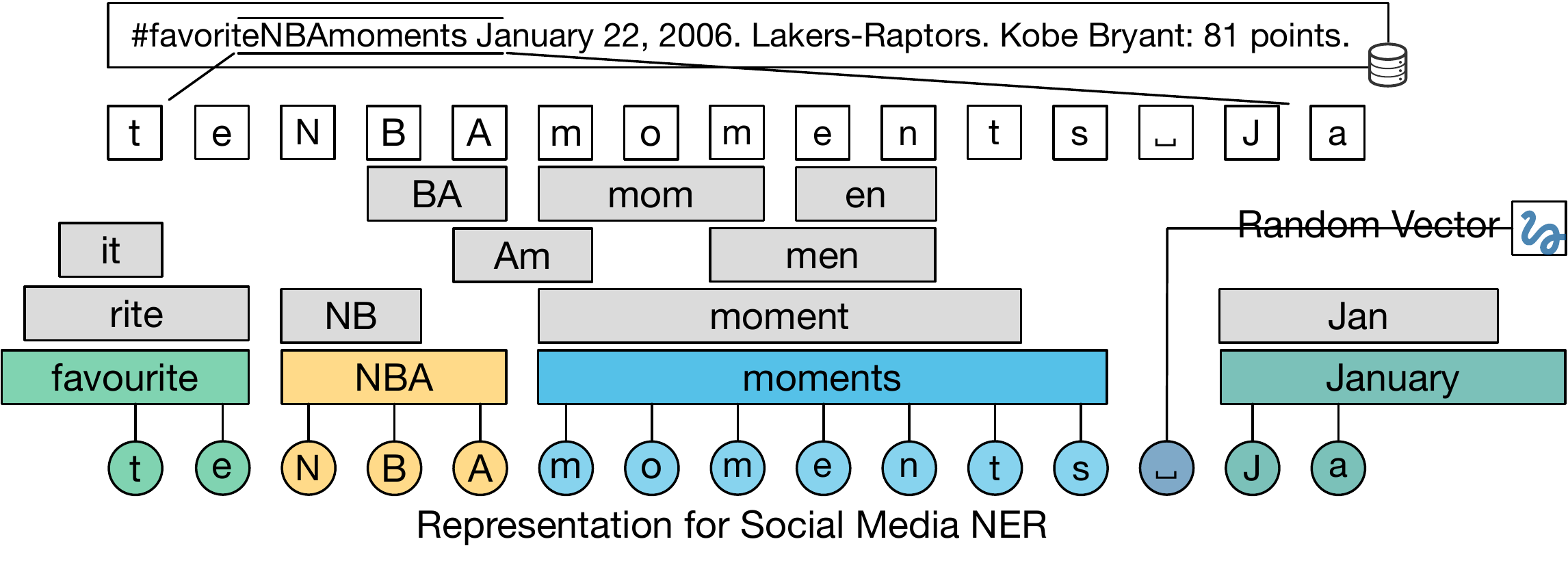}
  \vspace{-0.7cm}
  \caption{We used string matching to align the dictionary to the raw text for embedding learning.}
  \label{fig:string_match}
  \vspace{-0.4cm}
\end{figure}

\begin{figure*}[t]
  \centering
  \includegraphics[width=\linewidth]{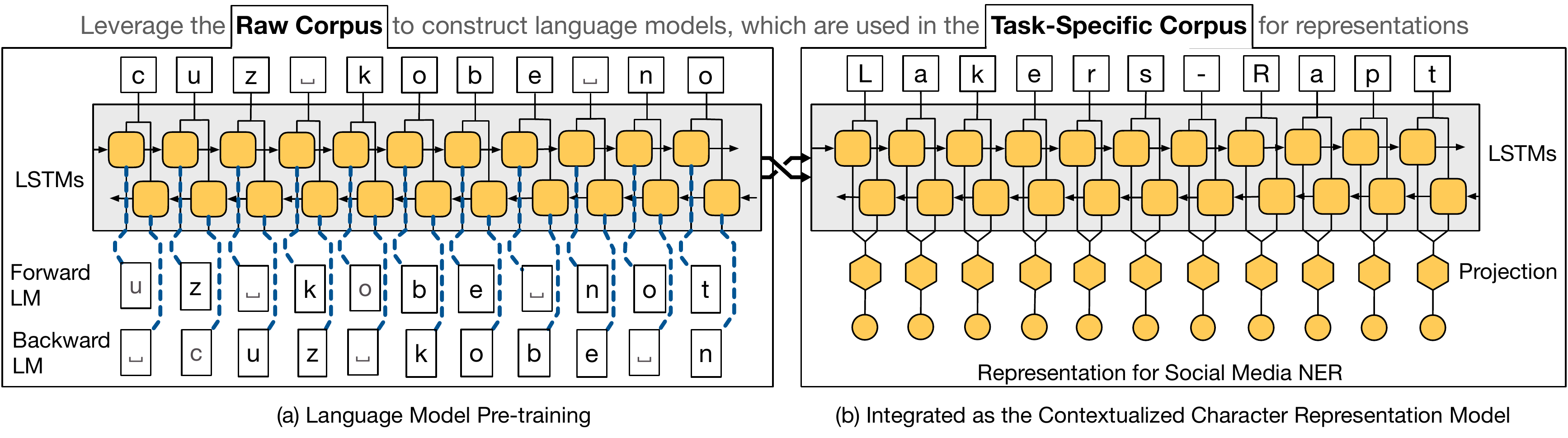}
  \vspace{-0.7cm}
  \caption{Contextualized character representation is constructed by pre-trained language models. Specifically, as in (a), the neural models are trained by predicting the afterwards / previous words; as in (b), these models are integrated into the downstream task to provide contextualized character representations.}
  \label{fig:lm}
  \vspace{-0.3cm}
\end{figure*}

\smallsection{String Matching-based Embedding Alignment}
To further reduce the reliance on pre-processing, we design a string matching-based approach to build the alignment in a tokenization-free manner. 
As shown in Figure~\ref{fig:string_match}, given the input character sequence, we first find all possible word matches from a given dictionary.
It is very likely that some characters will be matched to multiple words.
For example, the character ``N'' is matched to two words, ``NB'' and ``NBA''.
Therefore, for alignment purpose, we need to identify the most suitable match among these words.
We decide to choose the word that has the highest inverse document frequency (IDF) because words of higher IDF are more specific~\cite{sparck1972statistical}.
Particularly, we first sort all possible matches decreasingly by their IDFs.
Every time, we retrieve the match of the highest IDF for each character.
In this way, we give the priority to more specific and informative words.
As illustrated in Figure~\ref{fig:string_match}, this strategy can correctly identify the suitable match in most cases. 
After we finalize the matched word for each character, their word embedding vectors are used as representations for corresponding characters.

It is worth mentioning that this alignment process can be implemented in an efficient way.
To be more specific, by using the union-find set, the time complexity of this process can be bounded by $O(T\alpha(T))$, where $T$ is the input sequence length and $\alpha(\cdot)$ is the inverse Ackermann function. 
This function has a value $\alpha(T)<5$ for any value of $T$ that can be written in this physical universe, therefore the alignment process takes place in a linear time.
Moreover, the string matching-based alignment has a slight advantage over the tokenization-based alignment -- it requires no human endeavor to build a tokenizer.

\subsection{Contextualized Char Representation}

Contextualized representation learning at the word level has been widely adapted in state-of-the-art sequence labeling models.
They rely on bidirectional neural language models to capture the context information before and after a certain word.
This is a perfect supplementary to the context-agnostic information contained in word embedding.
Therefore, we design a contextualized \textit{character} representation model.
We first train bidirectional character-level language models with a large raw corpus (in Figure~\ref{fig:lm} (a)), and then integrate them into the \our framework (in Figure~\ref{fig:lm} (b)).

We present the details of character-level language modeling and the integration as follows. 

\smallsection{Character-Level Language Modeling}
As shown in Figure~\ref{fig:lm}, the bi-directional character-level language model contains two character-level language models.
Character-level language modeling aims to model the probability distribution of the character sequence. 
Typically, the probability of the sequence $\{x_1, \cdots, x_T\}$ is defined in a ``forward'' manner: $p(x_1, \cdots, x_T) = \prod_{t=1}^T p(x_t | x_1, \cdots, x_{t-1})$.
As in Figure~\ref{fig:lm}, we first map the input sequence $X$ to a list of character embedding vectors and pass them into a recurrent neural network, whose output is referred as $\h_t$. 
Then, the probability $p(x_t | x_1, \cdots, x_{t-1})$ is calculated using the softmax function.
The the backward language model is the same as the forward language model, except that it decomposes the probability of the sequence $\{x_1, \cdots, x_T\}$ as $p(x_1, \cdots, x_T) = \prod_{t=1}^T p(x_t | x_{t+1}, \cdots, x_T)$.
Its output for character $x_t$ is denoted as $\h_t^{r}$.
Both language models use the negative log-likelihood as the training objective. 

\smallsection{Language Model Integration}
Using the bidirectional character-level language models, we construct contextualized representations for each character.
Due to the complexity of natural language, large dimensions of $\h_t$ and $\h_t^r$ are usually required in language models, which could be too large for the NER task.
To avoid over-fitting, we add a linear transformation to project $\h_t$ and $\h_t^r$ to a lower dimension.
In details, we use $\r_t =W_{cr} \cdot [\h_t, \h_t^{r}] + \b_{cr}$, where $W_{cr}$ and $\b_{cr}$ are parameters to learn during the training of NER.
The output $\r_t$ is the contextualized representation for the character $x_t$, and serves as a part of its character representation.


\section{Experiment}
\label{sec:benchmark}


\subsection{Benchmark Datasets}

Here, to compare the model performance in a raw-to-end manner, we first convert word-level annotations to character-level annotations. 
Specifically, typical NER benchmark datasets are annotated after pre-processing, which includes but is not limited to adding/removing whitespaces by tokenization, spelling changes due to normalization, and even language changes using translation.
Consequently, this process is hard to be reverted without the original texts, and we choose two benchmark datasets, as follows, with the raw texts. 
\begin{itemize}[leftmargin=*,nosep]
\item \textbf{TNT}~\cite{lu2018visual} contains tweets from May 2016, January 2017 and June 2017, retrieved by using sports and social event related keywords. It is annotated with four types of entities(\texttt{PER}, \texttt{LOC}, \texttt{ORG} and \texttt{MISC}).

\item \textbf{BTC}~\cite{derczynski2016broad} contains tweets from various periods and regions, related to diverse topics. It is annotated with three types of entities(\texttt{PER}, \texttt{LOC} and \texttt{ORG}).

\end{itemize}

For both datasets, we adopt their recommended data split for training, development and testing sets, and summarize their statistics in Table~\ref{tab:stat}.
As to more details about the label conversion, please find the details in Appendix~\ref{app:untokenize}, which describes the rules we applied to map the pre-processed words back to the original raw texts and export character-level annotations. 


\begin{table}[t]
\begin{center}
\caption{Dataset Statistics of TNT and BTC.}
\vspace{-0.2cm}
\label{tab:stat}
\scalebox{0.8}{
\begin{tabularx}{1.2\columnwidth}{r *{6}{Y}}
\toprule
\multirow{2}{*}{Statistics} & \multicolumn{3}{c}{TNT} & \multicolumn{3}{c}{BTC} \\\cmidrule{2-7}
& Train & Dev & Test & Train & Dev & Test \\\midrule
\# Sent & 4290 & 1432 & 1459 & 6261 & 999 & 1998 \\
\# Token & 	69K & 23K & 23K & 98K & 15K & 35K \\
\# Char & 327K & 108K & 109K & 541K & 75K & 190K \\
\# \texttt{PER} & 3053 & 1016 & 991 & 3031 & 706 & 1600\\
\# \texttt{LOC} & 978 & 350 & 307 & 1981 & 151 & 601\\
\# \texttt{ORG} & 2617 & 889 & 964 & 2255 & 270 & 792 \\
\# \texttt{MISC} & 876 & 267 & 264 & \diagbox{}{} & \diagbox{}{} & \diagbox{}{} \\
\bottomrule
\end{tabularx}
}
\end{center}
\vspace{-0.5cm}
\end{table}

\subsection{Baseline Methods}
We pair state-of-the-art (Twitter) NER models with three popular pre-processing methods, which forms \emph{twelve baselines}.
Specifically, the four NER models we considered are:
\begin{enumerate}[nosep,leftmargin=*]
    \item \textbf{TwitterNER}~\cite{mishra2016_wnut_ner} is specifically designed for tweets. It relies on not only handcrafted features, but also domain-specific resources.
    \item \textbf{LSTM-CNNs-CRF}~\cite{ma2016end} does not rely on feature engineering and constructs the sequence labeling in an end-to-end manner. 
    \item \textbf{LM-LSTM-CRF}~\cite{liu2018empower} is a sequence labeling model, which integrates language models without pre-training and uses them to construct representations.
    \item \textbf{Flair}~\cite{akbik2018contextual} leverages language models to construct contextualized word representations and achieves the state-of-the-art performance on English NER for news corpus. 
\end{enumerate}
For pre-processing, we considered three tokenizers:
\begin{enumerate}[leftmargin=*,nosep]
    \item \textbf{spaCy}~\cite{spacy2} employs the NLP package spaCy for tokenization, which features efficiency and effectiveness\footnote{The version 2.0.18 release is used.}. 
    \item \textbf{Stanford CoreNLP}~\cite{manning-EtAl:2014:P14-5} leverages the widely used Stanford CoreNLP~\cite{manning-EtAl:2014:P14-5} as tokenizer\footnote{The version 3.9.2 release is used.}.
    \item \textbf{NLTK}~\cite{bird2008multidisciplinary} is a leading platform for building NLP pipelines and we only used its tokenization module here\footnote{The version 3.4 release is used.}.
\end{enumerate}
Since some of these pre-processing methods are not non-destructive, special handling is needed to convert entities to the original texts. 

\subsection{Training and Implementation Details}

In our experiments, each neural model was trained with one Nvidia V100 GPU or one Nvidia 1080 Ti GPU; other models were trained on a server with 20 cores of Intel Xeon CPU E5-2680 v2 @ 2.80GHz.
Our implementations would be released as a flexible framework, and more details about model training are described as below. 

\smallsection{Language models}
Following the previous work~\cite{akbik2018contextual}, we use one-layer LSTMs with 2048 hidden states in each direction.
The batch size is set to 128, and the training is conducted with backpropagation through time (BPTT) with a window size of 256.
We adopted a two-level encoding to better handle the noisy input, in which each character would be encoded as a type vector (\eg, number and lower case) and a character vector, and the concatenation of these two vectors would be passed to the LSTMs. 
We use Adam~\cite{kingma2014adam,Liu2019Radam} as the optimizer with default hyperparameters.
Dropout is applied with the probability of 0.01 and the gradient is clipped at 5. 
During training, we would anneal the learning rate by 0.1 if the loss stops for decreasing for three epoch, and halt the training when negligible gains were observed or after one week. 
We collect a large raw tweet corpus for embedding learning and language model training, specifically, we collect tweets with geo-location tag in the United States from January 1 to December 31 in 2015.
In total, we collect more than 700 million tweets consisting of more than 52 billion characters.
With this dataset, the forward language model achieves the character-level perplexity of 3.32 on the training corpus and the backward language model achieves 3.30.  

\begin{table}[t]
\begin{center}
\caption{Performance Comparison of Sequence Labeling Models and Tokenizers on Twitter NER Benchmarks. The reported numbers are \fl score. }
\vspace{-0.3cm}
\label{tab:all}
\scalebox{0.83}{
\begin{tabularx}{1.2\columnwidth}{c l *{2}{Y}}
\toprule
Tokenizer & Methods & TNT & BTC \\\midrule\midrule

\multirow{4}{*}{\begin{tabular}[c]{@{}c@{}} NLTK  \end{tabular}}

  &   {\small TwitterNER  } & 73.41 & 64.40\\
  &   {\small LSTM-CNN-CRF} & 80.25 & 66.48\\
  &   {\small LM-LSTM-CRF } & 80.85 & 67.73\\
  &   {\small Flair       } & \textbf{83.26} & \textbf{68.33} \\\midrule
  
\multirow{4}{*}{\begin{tabular}[c]{@{}c@{}}spaCy\end{tabular}}

  &   {\small TwitterNER  } & 70.40 & 37.21\\
  &   {\small LSTM-CNN-CRF} & 71.66 & 29.04 \\
  &   {\small LM-LSTM-CRF } & 72.25 & 31.17 \\
  &   {\small Flair       } & 73.49 & 33.20 \\\midrule

\multirow{4}{*}{\begin{tabular}[c]{@{}c@{}}Stanford \\CoreNLP \end{tabular}}

  &   {\small TwitterNER  } & 70.18 & 38.01 \\
  &   {\small LSTM-CNN-CRF} & 64.22 & 29.71 \\
  &   {\small LM-LSTM-CRF } & 63.14 & 28.97 \\
  &   {\small Flair       } & 65.58 & 32.39 \\\midrule\midrule

\multicolumn{2}{c}{ \our(Match)} & \textbf{86.96} & \textbf{74.98} \\
\multicolumn{2}{c}{ \our(NLTK)} & 85.59 & 74.39 \\

\bottomrule
\end{tabularx}
}
\end{center}
\vspace{-0.3cm}
\end{table}

\smallsection{Embedding models}
We build the embedding dictionary by following the previous work~\cite{liu2015mining,shangautomated}, a filter is further applied to remove words with the frequency less then 30. 
The embedding dimension is set to 100, the window size is set to 6 and the number of negative samples is set to 25.
We integrate the collected raw tweets corpus and the wiki dump\footnote{https://dumps.wikimedia.org/enwiki/latest/enwiki-latest-pages-articles.xml.bz2} for the model training, and optimization is conducted with SGD and stopped after 10 epochs.
The resulting embedding model contains about 1 million tokens and would be used in both our method and baseline methods for a fair comparison. 

\smallsection{Named Entity Recognition models}
We select most hyperparameters based on previous work~\cite{reimers2017optimal}, and adjust the optimizer hyperparameters based on its performance on the development set.
Specifically, the outputs of language models are projected to 100 dimension, the decoding part is equipped with 256 hidden states in each direction, the optimization would be conducted with Nadam~\cite{dozat2016incorporating}, and the gradient is clipped at 1. 
During the training, dropouts with the probability 0.5 are applied to every layer, and the output of all representation modules would be further randomly dropped with the probability of 0.1.
Since these two NER corpora are relative small, we would conduct the training of the final model on both the training set and the development set. 
The same strategy is also adopted for the training of baseline models. 

\subsection{Performance Comparison}

We summarize the performance in Table~\ref{tab:all}. 
Specifically, we consider two variants of the proposed method: \our (NLTK) uses NLTK-based embedding alignment and \our (Match) uses string matching-based embedding alignment. 
We can observe a significant performance improvement of \our over baseline methods (by $3.70$ and $6.65$ absolute \fl gain on TNT and BTC respectively), which verifies the effectiveness of our proposed framework. 
Further analyses are discussed as below.  


\begin{table}[t]
\begin{center}
\caption{Different word embeddings use different tokenizers. Even for the same word embedding,  different words could be tokenized differently (\eg, ``There's'' and ``That's'' as below). Intuitively, inappropriate tokenization results result in sub-optimal performance. }
\vspace{-0.3cm}
\label{tab:embed}
\scalebox{0.67}{
\begin{tabularx}{1.45\columnwidth}{c ll}
\toprule
\diagbox{Method}{Raw Text} & \texttt{There's} ... & \texttt{That's} ... \\\midrule
GloVe\footnotemark & \texttt{| There | 's |} ... & \texttt{| That | 's |} ...\\
fastText\footnotemark & \texttt{| There | 's |} ... & \texttt{| That's |} ... \\
Word2Vec\footnotemark & \texttt{| There's |} ... & \texttt{| That's |} ... \\
\bottomrule
\end{tabularx}
}
\end{center}
\vspace{-0.4cm}
\end{table}
\addtocounter{footnote}{-3} 
\stepcounter{footnote}\footnotetext{\url{http://nlp.stanford.edu/data/glove.840B.300d.zip}}
\stepcounter{footnote}\footnotetext{\url{https://dl.fbaipublicfiles.com/fasttext/vectors-english/crawl-300d-2M.vec.zip}}
\stepcounter{footnote}\footnotetext{\url{https://github.com/loretoparisi/word2vec-twitter}}

\smallsection{Comparing String Matching with NLTK}
We noticed that models with string matching-based alignment achieves similar and even better performance than models with NLTK-based alignment. 
We conduct further analysis to get more insights about this phenomenon.
As shown in Table~\ref{tab:embed}, different pre-processing rules are applied in different embedding learning methods, and thus lead to different dictionaries.
In previous NER methods or the toknenization induced word embedding approach, a simple Lookup Table query would be used to align the embedding dictionary with the pre-processed word sequence.
This practice may lead to a non-optimal alignment, when the pre-processing fails to meet some properties of the pre-trained embedding.
For example, if the tokenizer used for NER input didn't separate ``That'' and ``'s'' when using the pre-trained GloVe embedding~\cite{pennington2014glove}, the input ``That's'' would be mapped to the ``unknown token''. 
At the same time, our string matching approach is self-adaptive and has the ability to align different embedding individually. 
Therefore, by leveraging our proposed string matching approach, \our has the potential to be tokenization-free and can be constructed with less human endeavor. 


\begin{table}[t]
\begin{center}
\caption{Model Performance with Perfect Tokenization ($\overline{\Delta}$ is the average performance difference to those with three system tokenizers in Table~\ref{tab:all}).}
\vspace{-0.3cm}
\label{tab:two_step}
\scalebox{0.83}{
\begin{tabularx}{1.2\columnwidth}{r *{4}{Y}}
\toprule
&   \multicolumn{2}{c}{TNT}  &   \multicolumn{2}{c}{BTC} \\\cmidrule{2-3}\cmidrule{4-5}
& \fl & $\overline{\Delta}$ & \fl & $\overline{\Delta}$ \\\midrule
TwitterNER      &  73.24 & 1.91 & 65.67 & 19.13     \\
LSTM-CNN-CRF    &  81.74 & 9.70 & 66.48 & 25.02     \\
LM-LSTM-CRF     &  81.91 & 9.83 & 69.29 & 26.67        \\
Flair           &  84.34 & 10.23 & 71.41 & 26.77        \\
\bottomrule
\end{tabularx}
}
\end{center}
\vspace{-0.5cm}
\end{table}

\smallsection{Importance of Pre-processing}
One can easily observe that pre-processing plays a crucial role on the performance.
For example, switching the tokenizer from NLTK to spaCy or Stanford CoreNLP leads to a drop of more than 20 absolute \fl points. 
We further conduct experiments to verify our intuition to conduct NER in the raw-to-end manner. 
Specifically, we construct models with gold-tokenized training data and evaluate their performance with NLTK, which achieves the best performance among all three tokenizers. 
The results summarized in Table~\ref{tab:two_step} show that, after replacing NLTK-tokenized training data with gold-tokenized training data, \fl improves significantly. 
It verifies our intuition to build the raw-to-end framework.


\begin{table}[t]
\begin{center}
\caption{Ablation Study on Representation Modules.}
\vspace{-0.3cm}
\label{tab:ablation}
\begin{tabularx}{\columnwidth}{l *{3}{Y}}
\toprule
Method & TNT & BTC \\\midrule
{ \our (Match)}  & 86.96 & 74.98 \\ 
{ -- language model} & 80.72 & 63.71 \\
{ -- string match} & 75.28 & 59.35 \\
\bottomrule
\end{tabularx}
\end{center}
\vspace{-0.5cm}
\end{table}


\smallsection{Ablation Study}
As shown in Table~\ref{tab:ablation}, we conduct experiments with our best performing model (\ie, \our (Match)) to study the effect of the two representation modules. 
The first (referred as ``-- language model'') excludes the contextualized character representation module from \our, the second (referred as ``-- string match'') replaces the proposed two modules with static character embedding. 
The results demonstrate the effectiveness of both proposed representation modules -- significant performance improvements can be observed by adapting either of these two approaches. 


\begin{table*}[t]
\begin{center}
\caption{Case study on the TNT dataset. Incorrect outputs are marked as red and bold.}
\vspace{-0.3cm}
\label{tab:case_study}
\scalebox{0.74}{
\begin{tabularx}{1.34\linewidth}{c lll}
\toprule
 & 
\begin{tabular}[c]{@{}l@{}}
LeBron named to 10th All-\#NBA \\First Team
\end{tabular} & \, &
\begin{tabular}[c]{@{}l@{}}
Moving on up: @GeeksOUT and @FLAMECON\\ are now on display in Times Square.
\end{tabular}
\\\midrule

\begin{tabular}[c]{@{}c@{}}
\textbf{\small Flair (NLTK)}
\end{tabular}
& 
\begin{tabular}[c]{@{}l@{}}
$<$LeBron, PER$>$ $<$NBA, ORG$>$ 
\end{tabular}
& \, &
\begin{tabular}[c]{@{}l@{}}
\wrong{$<$GeeksOUT, PER$>$}
\wrong{$<$FLAMECON, PER$>$}
$<$Times Squere, LOC$>$ 
\end{tabular}
\\\midrule

\textbf{\small Flair (spaCy)} & 
\begin{tabular}[c]{@{}l@{}}
$<$LeBron, PER$>$ \wrong{$<$All-\#NBA, O$>$}
\end{tabular}
& \, &
\begin{tabular}[c]{@{}l@{}}
\wrong{$<$@GeeksOUT, PER$>$}
\wrong{$<$@FLAMECON, ORG$>$}
$<$Times Squere, LOC$>$ 
\end{tabular}
\\\midrule

\begin{tabular}[c]{@{}l@{}}
\textbf{\small Flair (Stanford} \\\textbf{\small CoreNLP)}
\end{tabular}
& 
\begin{tabular}[c]{@{}l@{}}
$<$LeBron, PER$>$ \wrong{$<$\#NBA, ORG$>$}
\end{tabular}
& \, &
\begin{tabular}[c]{@{}l@{}}
\wrong{$<$@GeeksOUT, PER$>$}
\wrong{$<$@FLAMECON, PER$>$}
$<$Times Squere, LOC$>$ 
\end{tabular}
\\\midrule

\begin{tabular}[c]{@{}c@{}}
\textbf{\small Neural-Char-CRF} 
\end{tabular}
& 
\begin{tabular}[c]{@{}l@{}}
$<$LeBron, PER$>$
$<$NBA, ORG$>$ 
\end{tabular}
& \, &
\begin{tabular}[c]{@{}l@{}}
$<$GeeksOUT, ORG$>$
$<$FLAMECON, ORG$>$
$<$Times Squere, LOC$>$ 
\end{tabular}
\\
\bottomrule
\end{tabularx}
}
\vspace{-0.3cm}
\end{center}
\end{table*}

\subsection{Case Studies}
Table~\ref{tab:case_study} shows the output of \our and a strong baseline \method{Flair} on the TNT dataset. 
As for the first sentence, since both spaCy and Stanford CoreNLP fail to detect the word boundary of ``NBA'', methods that rely on tokenization results cannot identify this entity successfully. 
At the same time, the second sentence exhibits that \our can effectively leverage the context information.

\section{Related Work}
\label{sec:related}

\subsection{Named Entity Recognition}

Most NER systems, as mentioned before, are constructed as sequence labeling models.
Typically, in traditional methods, handcrafted features are leveraged to capture textual signals, and conditional random fields (CRF) are employed to model label dependencies~\cite{finkel2005incorporating,settles2004biomedical,leaman2008banner}.
Recent advances in neural models allow us to better represent natural language and freed domain experts from handcrafting features on many tasks.
\method{Bi-LSTM-CRF} leverages both word embedding and handcrafted features, combines neural networks with CRF and shows improvements over previous methods~\cite{huang2015bidirectional};
\method{LSTM-CNN} further incorporates CNN and illustrates the potential of capturing character-level signals~\cite{chiu2016named};
\method{LSTM-CRF} and \method{LSTM-CNN-CRF} are proposed to get rid of hand-crafted features and demonstrate the feasibility to fully rely on representation learning to capture the textual signals~\cite{lample2016neural,ma2016end}.
Meanwhile, \method{CharNER} was proposed to make initial predictions at the character-level, however, this method still relies on tokenization to further regularizes and refines model outputs to ensure that the detected word boundaries is maintained in its final predictions, and thus it can be viewed as a variant of the word-level model~\cite{kuru2016charner}. 
More recently, language modelings are noticed to be dramatically effective as the representation module for NER~\cite{peters2017semi,peters2018deep,liu2018efficient,liu2018empower,akbik2018contextual,liu2019arabic}.
Comparing these methods, it is noticed that most improvements are brought by new representation techniques, which allows us to construct the model in a more data-driven manner.
Despite the effectiveness of these methods, as discussed before, all of them rely on preprocessing components and would have deteriorated performance on noisy texts like social media. 
In this paper, we propose to conduct NER in a raw-to-end manner.

\subsection{Word Embedding}

Most word embedding methods are based on the distributional hypothesis, \ie, ``a word is characterized by the company it keeps''~\cite{harris1954distributional}, and learn word representations by analyzing their contexts.
Unlike the previous work~\cite{bengio2003neural,hinton1986learning}, word2vec~\cite{mikolov2013efficient} leverages the hierarchical softmax and the negative sampling, thus can scale up to extensive corpora.
Recent work shows that word embedding could cover textual information of various levels~\cite{artetxe2018uncovering}.
It has been shown that, by properly handling such embedding, the model performance can be boosted significantly~\cite{liu2019arabic, lin2019reliability}. 
Instead of trusting the pre-processing results, we develop two strategies to build robust word embedding alignments that suffer less from pre-processing errors.

\subsection{Contextualized Representations}

By leveraging pre-trained machine translation models, \method{CoVe} constructs contextualized word representations~\cite{mccann2017learned}.
After that, \method{ELMo} replaces translation with language modeling, which does not require annotations and has nearly unlimited corpora~\cite{peters2018deep}.
It demonstrates significant improvements on various NLP tasks. 
A more comprehensive comparison shows that, comparing to machine translation~\cite{zhang2018language}, language modeling is more effective as the pre-training task, even after limiting the size of its training data. 
At the same time, lots of attentions have been attracted to leveraging language modeling to build sentence representations~\cite{howard2018universal,radford2018improving,devlin2018bert}.
Besides word-level language models, character-level language models have also been leveraged to construct the contextualized representation~\cite{liu2018empower,akbik2018contextual}. 
These models are designed to get the contextualized representation for a span of characters (i.e., words and sentences).
Here, we present a contextualized character representation model, which provides context information at the character-level.


\section{Conclusion and Future Work}
\label{sec:con}

In this paper, we study the Named Entity Recognition task in noisy, user-generated texts. 
Recognizing the importance of pre-processing, we propose a raw-to-end framework \our.
It takes raw input as character sequences and makes end-to-end predictions, thus relying less on pre-processing and suffering less from error propagation.
Two novel representation learning modules are tailored to better capture the textual signals.
Empirical results on the two tweets datasets demonstrate the superior performance of the proposed \our method.
Performance analysis, ablation study and case studies further verify our intuitions. 
In the future work, 
We plan to apply our method to other tasks, domains and languages.

\bibliography{cited}
\bibliographystyle{acl_natbib}

\appendix
\clearpage

\section{Benchmark Annotation Conversion}
\label{app:untokenize}

In our implementations, there are three stages for recovering the character-level annotations from word-level annotations. 
First, we applied several pre-processing pipelines to regularize the raw text (include html unescape and unicode normalize). 
Comparing their inputs and outputs, we can build a dictionary from the raw character sequences to the processed character sequences. 
Secondly, we revert this dictionary, remove all spaces in the original texts and directly concatenate all words in the tokenized sequence. 
Then we try to align these two strings through the dictionary built in the last step, and record this alignment as a index mapping.
In the end, we would calculate entity positions based on the index mapping, and generate character-level annotations based on the BIOES schema.  

\section{Pipeline Implementation Details}
\label{app:preprocess}

\begin{figure}[ht]
  \centering
  \includegraphics[width=0.7\linewidth]{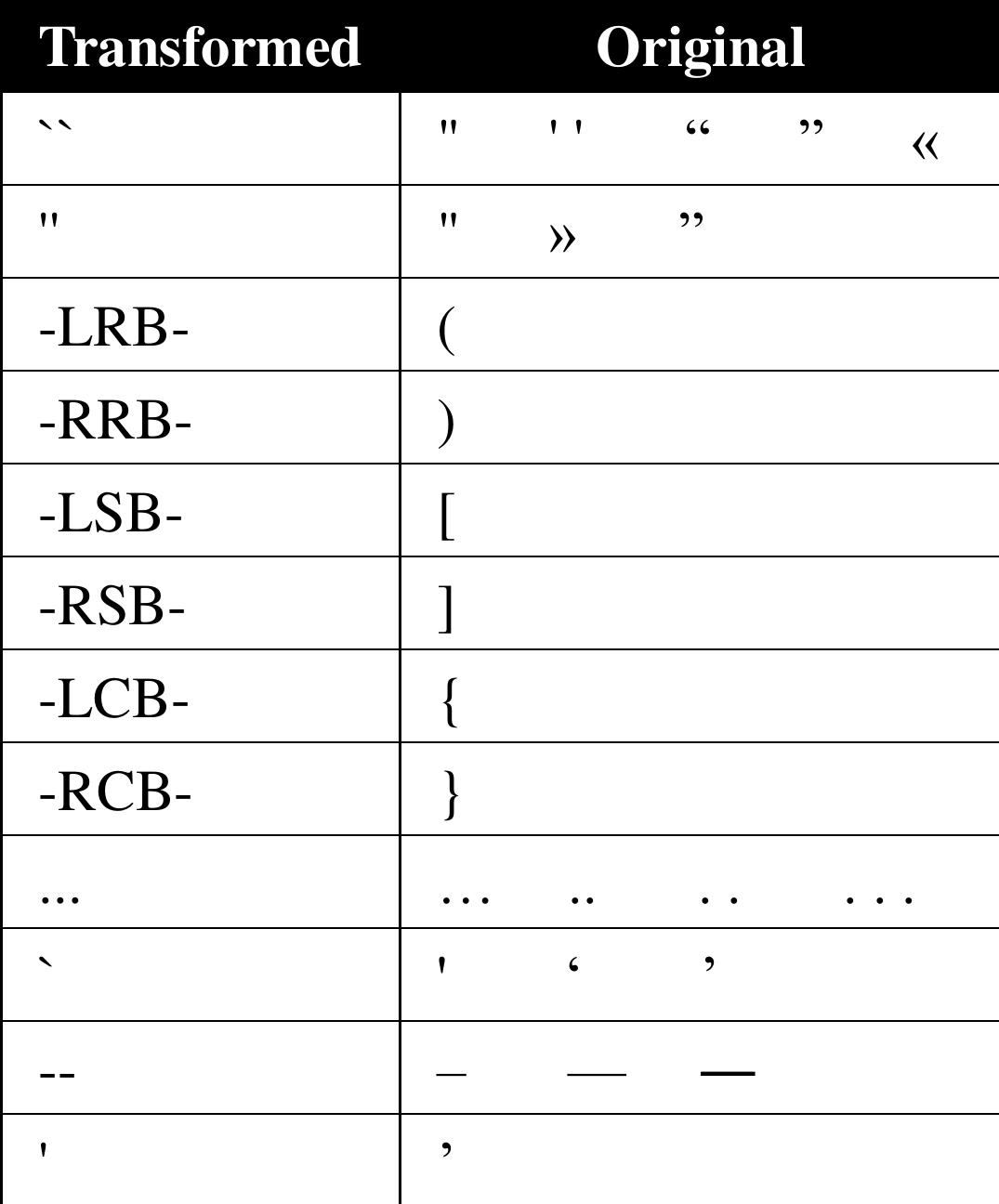}
  \caption{Table reflects the mapping from the original characters to their transformed characters by Stanford Tokenizer. In this process, different characters can be transformed to a same sequence of characters, and different sequences of characters may come from a same character.}
  \label{fig:token_map}
\end{figure}

Stanford and spaCy tokenizer are used to detect the boundaries in the text by taking in raw text input and outputing a list of tokenized words. However, Stanford tokenizer applies transformation on certain characters, which makes the tokenized result different from the original text. For example, `\textbf{(}' is transformed into `\textbf{-LRB-}', and `\textbf{. .}' is transformed into `\textbf{...}' (quotes added for clarity). This causes problems in boundary detection as the tokenized words and original text do not match. We looked through all the data in \textbf{TNT} and \textbf{BTC} and collected a list of transformed words and their original text, see Table~\ref{fig:token_map}. The transferred words are mapped back to find the correct boundaries. Its worthwhile to mention that, we only map the tokenized words back to detect boundaries in original text and to compare predicted entities with gold-standard. When it comes to training the model and making predictions, we still use tokenized words without mapping. \\
Different text may be transformed to the same tokenized word, so we find the best mapping such that between all consecutive boundaries detected, only spaces or untokenizable characters remain.

\end{document}